\documentclass[10pt,twocolumn,letterpaper]{article}

\usepackage[pagenumbers]{cvpr} 

\newcommand{\Ourname}{Anomaly-Aware CLIP\xspace}
\newcommand{\ourname}{AA-CLIP\xspace}

\usepackage{multirow}
\usepackage{multicol} 
\usepackage{booktabs}
\usepackage{tabularx} 
\usepackage{graphicx} 
\usepackage{adjustbox} 
\usepackage{colortbl} 
\usepackage{caption} 
\usepackage{appendix}
\usepackage[dvipsnames]{xcolor} 
\colorlet{colorFst}{Green!25}       
\colorlet{colorSnd}{SpringGreen!45} 
\colorlet{colorTrd}{Yellow!30}      
\colorlet{colorLow}{darkgray!30}    
\colorlet{colorDeg}{Orange!0}      
\newcommand{\fs}{\cellcolor{colorFst}\bf}   
\newcommand{\nd}{\cellcolor{colorSnd}}      
\newcommand{\rd}{\cellcolor{colorTrd}}      

\usepackage{algorithm}
\usepackage{algpseudocode}

\definecolor{cvprblue}{rgb}{0.21,0.49,0.74}
\usepackage[pagebackref,breaklinks,colorlinks,allcolors=cvprblue]{hyperref}

\def\paperID{1380} 
\def\confName{CVPR}
\def\confYear{2025}

\title{AA-CLIP: Enhancing Zero-Shot Anomaly Detection via Anomaly-Aware CLIP}
\vspace{-1.5mm}

\author{Wenxin Ma$^{1,2}$\qquad
Xu Zhang$^{1,2}$\qquad 
Qingsong Yao$^{5}$ \qquad 
Fenghe Tang$^{1,2}$ \qquad 
Chenxu Wu$^{1,2}$ \\
Yingtai Li$^{1,2}$ \qquad 
Rui Yan$^{1,2}$ \qquad 
Zihang Jiang$^{1,2}$\footnotemark[1] \qquad 
S.Kevin Zhou$^{1,2,3,4}$\thanks{Corresponding author.}
\\
$^1$ School of Biomedical Engineering, Division of Life Sciences and Medicine, USTC\\
$^2$ MIRACLE Center, Suzhou Institute for Advance Research, USTC \\
$^3$  Key Laboratory of Intelligent Information Processing of CAS, ICT, CAS \\
$^4$ State Key Laboratory of Precision and Intelligent Chemistry, USTC \\
$^5$ Stanford University \\
{\tt\small wxma@mail.ustc.edu.cn jzh0103@ustc.edu.cn s.kevin.zhou@gmail.com}
}

\let\oldtwocolumn\twocolumn
\renewcommand\twocolumn[1][]{%
    \oldtwocolumn[{#1}{
    \begin{center}
    \captionsetup{type=figure}
    \vspace{-5mm}
    \includegraphics[width=\textwidth]{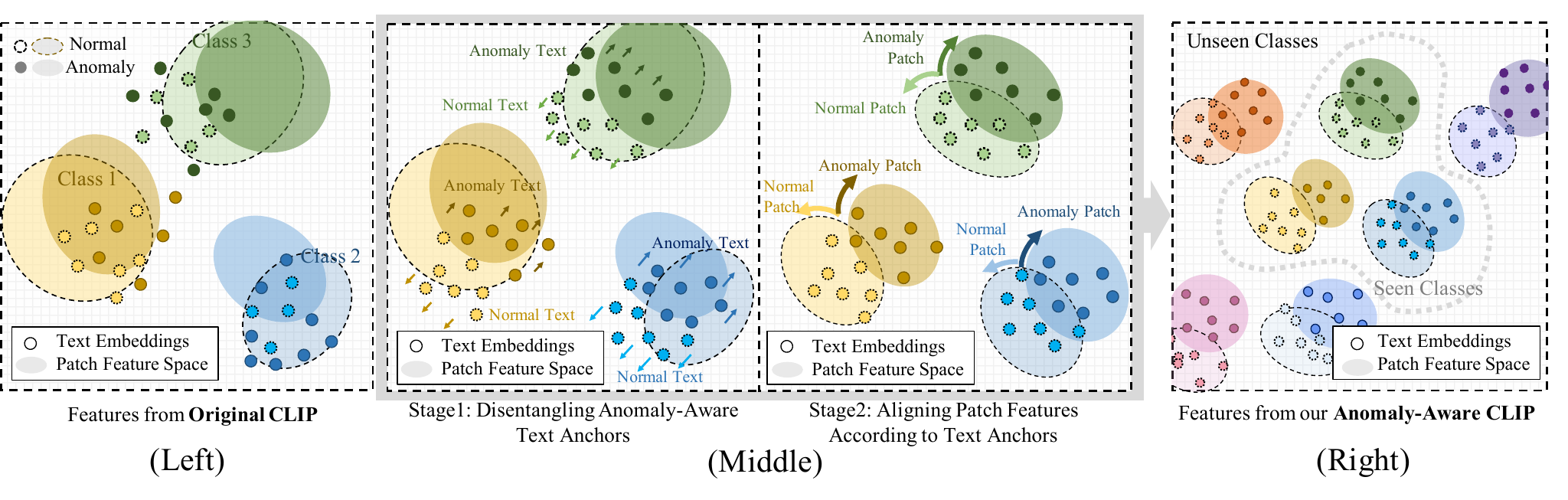}
    \vspace{-8mm}
    \captionof{figure}{\textbf{(Left) CLIP's anomaly unawareness:} Category-level image-text alignment in pre-training leads to CLIP's vague distinctions in anomaly/normal semantics and inaccurate patch-text alignment. \textbf{(Middle) Our two-stage adaptation strategy:} In Stage1, anomaly and normal text features are disentangled as anchors in text space; in Stage2, patch-level visual features are trained to align to these anchors, forming Anomaly-Aware CLIP. \textbf{(Right) Generalizable anomaly awareness:} Our method enables CLIP with generalizable anomaly awareness for both known and unseen classes.}
    \label{fig:teaser}
  \end{center}
    }]
}

\begin{document}
\maketitle

\begin{abstract}
Anomaly detection (AD) identifies outliers for applications like defect and lesion detection. While CLIP shows promise for zero-shot AD tasks due to its strong generalization capabilities, its inherent \textbf{Anomaly-Unawareness} leads to limited discrimination between normal and abnormal features. To address this problem, we propose \textbf{Anomaly-Aware CLIP} (AA-CLIP), which enhances CLIP's anomaly discrimination ability in both text and visual spaces while preserving its generalization capability. AA-CLIP is achieved through a straightforward yet effective two-stage approach: it first creates anomaly-aware text anchors to differentiate normal and abnormal semantics clearly, then aligns patch-level visual features with these anchors for precise anomaly localization. This two-stage strategy, with the help of residual adapters, gradually adapts CLIP in a controlled manner, achieving effective AD while maintaining CLIP's class knowledge. Extensive experiments validate AA-CLIP as a resource-efficient solution for zero-shot AD tasks, achieving state-of-the-art results in industrial and medical applications. The code is available at \url{https://github.com/Mwxinnn/AA-CLIP}. 
\end{abstract}  
\vspace{-2.5mm}
\section{Introduction}
\vspace{-1.5mm}
Anomaly detection (AD) involves modeling the distribution of a dataset to identify outliers, such as defects in industrial products~\cite{bergmann2019mvtec} or lesions in medical images~\cite{fernando2021deep}. Despite that previous AD frameworks~\cite{defard2021padim,deng2022anomaly,liu2023simplenet,kim2023sanflow,gudovskiy2022cflow,zhang2023destseg} effectively detect anomalies when sufficient labeled data is available for specific classes, their high resource demands often limit their generalization ability to novel and rare classes.  
This limitation is particularly challenging in real-world scenarios where collecting comprehensive labeled datasets for AD is often infeasible, necessitating the exploration of low-shot learning and transfer learning approaches.

Contrastive Language-Image Pretraining (CLIP) model has emerged as a promising solution, demonstrating remarkable generalization capabilities across various zero-shot tasks~\cite{radford2021learning, li2021align, li2022blip, li2023blip}. Building upon CLIP's success, several recent studies have adapted CLIP for few/zero-shot AD tasks by utilizing anomaly-related descriptions to guide the detection of anomalous regions.
Specifically, the vision encoder is trained to map anomaly images to visual features that align more closely with text features of abnormal descriptions than with those of normal descriptions~\cite{wang2022cris, liu2023clip, zhou2023zegclip, lin2023clip}.
Further works~\cite{Huang_2024_CVPR, cao2025adaclip, chen2023april, qu2024vcp} have focused on enhancing CLIP's patch-level feature representations to achieve better alignment with text features, resulting in improved anomaly localization performance.

These methods depend on text features that need to be anomaly-aware to effectively differentiate abnormalities. However, recent studies highlight CLIP’s limitations in fine-grained semantic perception and reasoning~\cite{tang2023lemons, jiang2023clip, paiss2023teaching, subramanian2022reclip, momeni2023verbs, khan2024figclip}. 
Upon exploring CLIP's texture features for AD, we observe that while CLIP’s text encoder effectively captures object-level information, it struggles to reliably distinguish between normal and abnormal semantics. As shown in conceptual visualization \cref{fig:teaser}(left) and sampled examples in \cref{fig:similarity}, CLIP has the intrinsic \textbf{Anomaly-Unawareness} problem: the overlap of normal and abnormal texture features hampers the precision of text-guided anomaly detection. We argue that making CLIP anomaly-aware — by establishing clearer distinctions between normal and abnormal semantics in the text space — is essential for guiding the vision encoder to precisely detect and localize anomalies.

This observation drives us to improve CLIP-based zero-shot AD through enhancing anomaly discrimination in text space, achieved with our method \textbf{\Ourname(\ourname)} — a CLIP model with anomaly-aware information encoded. \ourname is implemented through a novel two-stage adaptation approach.
In the first stage, \ourname adapts the text encoder with frozen visual encoder, creating ``anchors'' for anomaly-aware semantics within the text space for each trained class. As illustrated in \cref{fig:teaser}(middle), each class’s text features are disentangled to distinct anchors, with clear abnormality discrimination. Notably, this disentanglement also applies to novel, unseen classes, supporting effective zero-shot inference in AD tasks (refer to \cref{fig:teaser}(right)).
In the second stage, \ourname aligns patch-level visual features with these specially adapted texture anchors, guiding CLIP’s visual encoder to concentrate on anomaly-relevant regions. This two-stage approach ensures a focused and precise anomaly detection framework.
 
Importantly, as CLIP is extensively trained on massive data, to preserve its pre-trained knowledge, we utilize simple-structured Residual Adapters in both stages. This design enables a controlled adaptation of CLIP while enhancing its capability to handle fine-grained AD tasks without sacrificing its generalization ability. 

Our extensive experiments in both industrial and medical domains demonstrate that our straightforward approach equips CLIP with improved zero-shot AD ability, even in data-limited scenarios. By training with a minimal sample — such as one normal sample and one anomaly sample (2-shot) per class — and testing across unseen datasets, our method achieves zero-shot performance comparable to other CLIP-based AD techniques. With only 64-shot of each class seen in the training set, our method reaches state-of-the-art (SOTA) results in cross-dataset zero-shot testing, validating our method's ability to maximize the CLIP’s potential for AD with a minimal data requirement.

Our contributions are summarized as follows:
\begin{enumerate}
    \item \textit{Anomaly-Aware CLIP with enhanced and generalizable anomaly-discriminative ability.} We introduce \ourname which is more sensitive to anomalies sequentially in text and visual spaces, encoding anomaly-aware information into the original CLIP.
    \item \textit{Efficient adaptation using residual adapters.} We implement simple residual adapters to boost zero-shot anomaly detection performance without compromising the model’s generalization ability.
    \item \textit{SOTA performance with high training efficiency.} Our method achieves SOTA results across diverse datasets, showing robust anomaly detection capabilities even with limited training samples.
\end{enumerate}

\section{Related Work}
\vspace{-1.5mm}
\noindent\textbf{Traditional Anomaly Detection} in images involves modeling the normal data distribution to detect rare and diverse unexpected signals within visual data~~\cite{zong2018deep,  schlegl2017unsupervised, yao2021label}. Reconstruction-based~~\cite{yao2023one,  he2023diad,  lu2023hierarchical,  you2022adtr,  deng2022anomaly, Ma_2025_WACV},  augmentation-based~\cite{schlegl2017unsupervised,  liu2023simplenet,  you2022unified,  zhang2023destseg,  wang2021student} and discriminative~\cite{zong2018deep,  defard2021padim,  roth2022towards,  liu2023simplenet,  gudovskiy2022cflow,   kim2023sanflow} methods are typically used to facilitate better modeling. Despite the huge progress of traditional anomaly detection methods,  their effectiveness relies heavily on a well-modeled normal data distribution. Without sufficient normal data,  their ability to accurately detect anomalies is significantly reduced.

\vspace{1.2mm}
\noindent\textbf{CLIP}, trained on a vast amount of image-text data, leverages contrastive learning alongside powerful language models and visual feature encoders to capture robust concepts. This combination enables CLIP to achieve impressive zero-shot performance on image classification, as it can generalize well to new categories without requiring task-specific training~\cite{radford2021learning,  li2021align,  li2022blip,  li2023blip,  li2024graphadapter,  yang2023diffusion}. More recently, numerous studies~\cite{fang2021clip2video, mokady2021clipcap, cho2022fine, patashnik2021styleclip} have explored ways to transfer the knowledge embedded in CLIP models to a variety of downstream tasks, yielding promising results in fields like image captioning, image-text retrieval, and image generation. These efforts demonstrate CLIP's versatility and potential to drive advancements across diverse applications.

Despite the rapid advancements achieved by CLIP, numerous studies have highlighted persistent limitations in the features it extracts. While CLIP demonstrates strong generalization across various tasks, it often struggles to capture nuanced details and essential spatial relationships, which are crucial for tasks demanding precise boundary delineation and fine-grained feature extraction. This limitation results in suboptimal performance in downstream applications, especially that require high levels of detail, such as object detection, scene segmentation, or tasks in medical imaging~\cite{wang2022cris, liu2023clip, zhou2023zegclip, lin2023clip, gao2024clip, zhang2022tip, xu2023side, monsefi2024detailclip}. As a result, leveraging CLIP for fine-granular tasks frequently necessitates task-specific adaptations to bridge the gap between its generalized feature extraction and the precision required for specialized applications.

\vspace{1.2mm}
\noindent\textbf{CLIP-based Anomaly Detection}
There have been several efforts to leverage CLIP for AD tasks. One of the pioneering approaches, WinCLIP~\cite{jeong2023winclip}, proposes a method for extracting and aggregating visual features from multiple levels to align with text features, demonstrating the potential of CLIP in this context. Subsequent research investigates various adaptation methods to bridge the gap between natural domains and the AD domain, resulting in performance improvements. For instance,
~\cite{chen2023clip, huang2024adapting, chen2023april} focus on refining visual features by employing adapters to enhance patch-level visual representations. However, these approaches often rely on text embeddings from the original CLIP model as soft supervision and overlook a critical limitation of CLIP in AD: its unclearness in distinguishing between anomalous and normal semantics, particularly within the text encoder, resulting in suboptimal performance.
Other works have employed prompt-learning-based methods\cite{cao2025adaclip, zhou2023anomalyclip, qu2024vcp}, introducing learnable embeddings into the text encoder to better represent abnormality. However, the class information in CLIP can be damaged, potentially degrading generalization, especially in data-limited and zero-shot settings.

Different from previous methods, we are the first to investigate CLIP's inherent limitation in capturing anomaly-aware information, specifically in differentiating between normal and anomalous semantics in text prompts. Rather than relying solely on the original anomaly-unaware text embeddings or unaltered feature spaces, our method is able to refine the embeddings to actively incorporate anomaly-discriminative representations. 
\section{Method}
\subsection{Overview}
\subsubsection{Problem Formulation}
Zero-shot AD models are trained to identify anomalous samples whose categories may be unseen in the training dataset. Specifically, the model is expected to learn both normal and abnormal patterns that are shared across different classes given a training set $\mathcal{D}_{train}$ with normal or anomalous samples, in order to be capable of performing AD tasks on a series of different test datasets $\{\mathcal{D}_{test}^1, \mathcal{D}_{test}^2, ..., \mathcal{D}_{test}^{n}\}$, where each $\mathcal{D}_{test}^i$ is distinct from $\mathcal{D}_{train}$.
Image-level AD can be formally defined as a binary classification problem, where the model aims to classify samples $x\in \mathcal{D}$ as either normal ($y=0$) or anomalous ($y=1$). Anomaly segmentation extends this concept to pixel-level with mask $S$, aiming to identify anomalous regions by highlighting pixels associated with anomalies.

\subsubsection{Current Challenges}
\label{sec:current-challenges}
\paragraph{Anomaly Unawareness in CLIP:} The CLIP-based AD method classifies visual features as ``anomalies'' if they exhibit greater similarities to anomaly prompt embeddings than to normal prompt embeddings, thus requiring well-defined boundaries between these two kinds of prompts. 
However, in real applications, CLIP's text embeddings often lack the clear separability needed to reliably distinguish between normal and anomaly classes.

{\small
\begin{figure}[t]
    \centering
    \includegraphics[width=\linewidth]{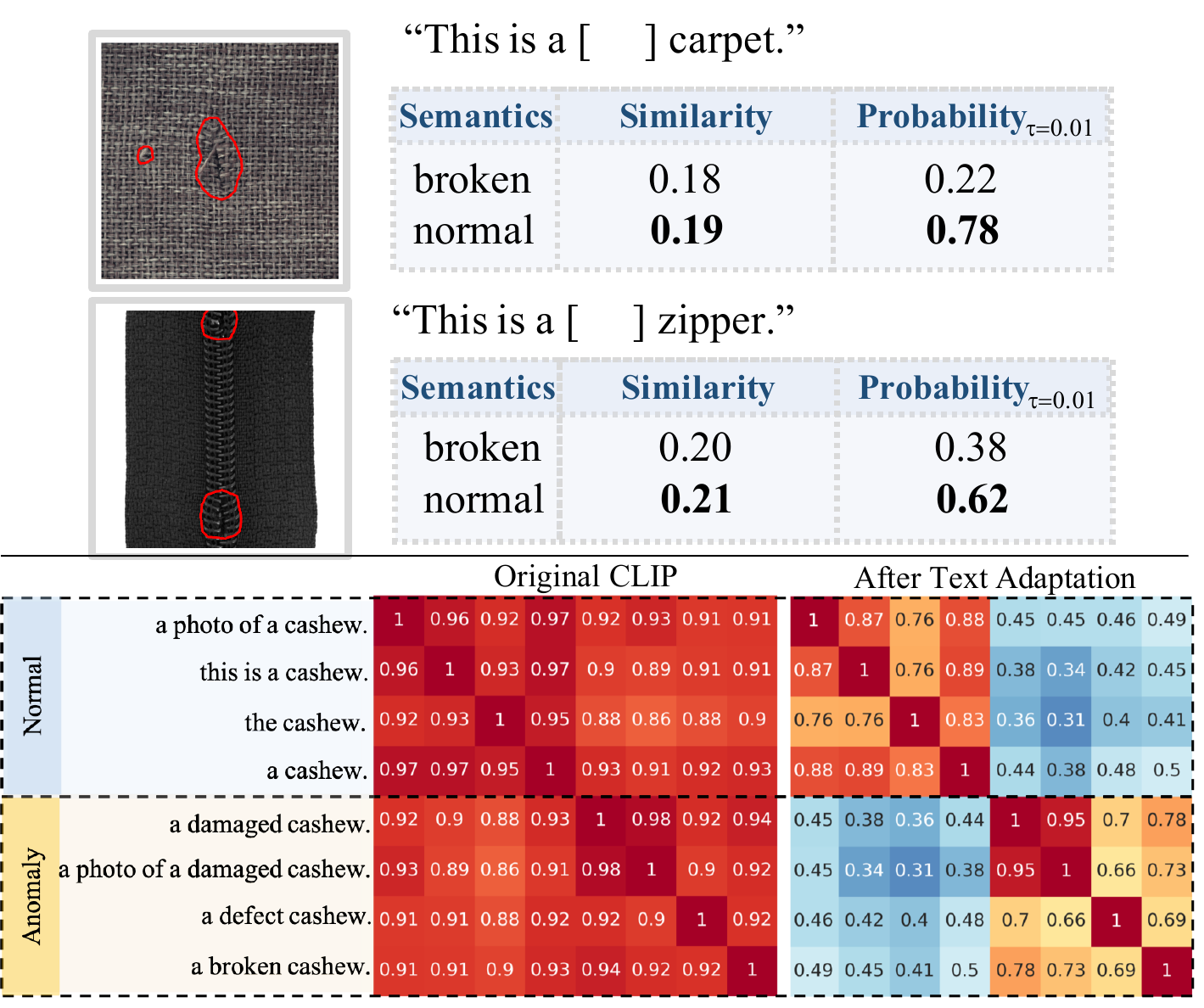}
    \vspace{-7.5mm}
    \caption{\textbf{(Top) Examples illustrating CLIP's Anomaly Unawareness.} 
    Despite the obvious anomalies present in the images, image features have higher similarities to normal descriptions, rather than anomaly descriptions, mistakenly. This problem is enlarged with a low temperature $\tau$. \textbf{(Bottom) Text Feature Similarity Heatmap among Normal and Anomaly Descriptions: Original CLIP vs. After Text Adaptation.} Red indicates high similarity. In original CLIP, normal features exhibit strong similarity with anomaly features, whereas text adaptation successfully separates them, clarifying the semantic distinctions between normal and anomaly descriptions.}
    \vspace{-2.5mm}
    \label{fig:similarity}
\end{figure}
}
We observe that, despite the visible defects in example images from the MVTec-AD~\cite{bergmann2019mvtec}, their features exhibit higher cosine similarity with ``normal'' prompts than with correct ``anomaly'' descriptions (see \cref{fig:similarity} (top)), indicating CLIP's inaccurate semantic understanding. Without adaptation, there persists a high similarity between the normal and abnormal text embeddings of a single class, as shown in \cref{fig:similarity} (bottom), suggesting a potential entanglement of normal and anomaly semantics within text space.  
We term this limitation \textbf{Anomaly Unawareness} and attribute it to the training process of CLIP: it is primarily trained on general, non-anomalous datasets and lacks specific guidance on defect detection. Consequently, it is challenging to rely on original CLIP embeddings to detect subtle or context-specific anomalies.
{\small
   \begin{figure}[t]

    \centering\includegraphics[width=\linewidth]{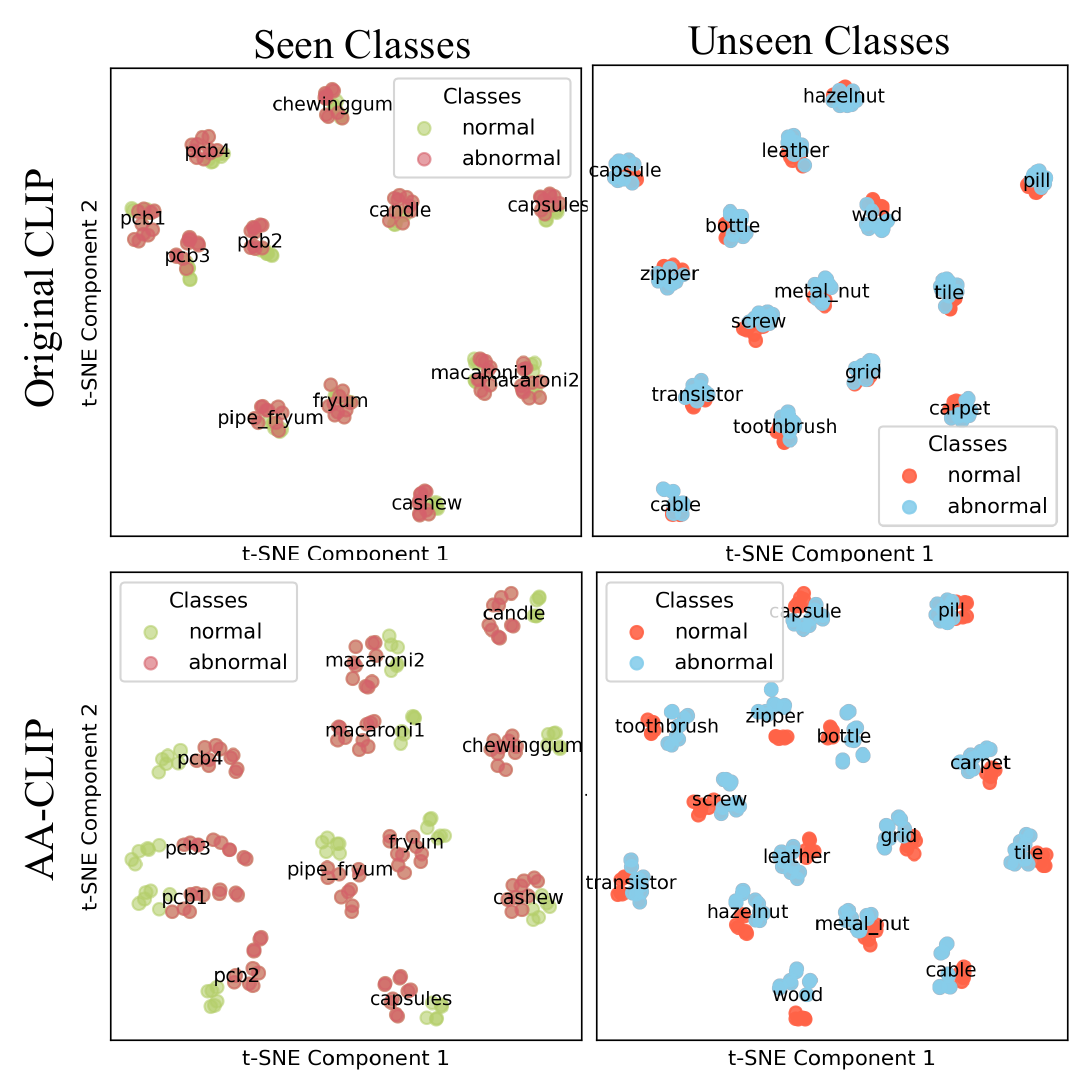}
    \vspace{-4.5mm}
       \caption{\textbf{t-SNE Visualization of Text Features from Original CLIP vs. AA-CLIP.} Each point represents a text feature encoded from a prompt. Original CLIP's normal and anomaly text features are intertwined, while our method effectively disentangles them. This disentanglement is generalizable to novel classes, validating the anomaly-awareness of our model.}
       \vspace{-3.5mm}
       \label{fig:tsne}
   \end{figure}
}
 
This issue remains evident across different categories in our t-SNE analysis: as shown in \cref{fig:tsne} (top), only subtle separations are observed within an object cluster, where text embeddings for both normal and abnormal semantics are intermixed. This entangled pattern may potentially lead to anomaly-unaware text-image alignment, which reinforces the necessity to \textit{adapt CLIP’s to enhance its ability of anomaly-awareness.} 

\vspace{-3.3mm}
\paragraph{Embedding Adaptation Dilemma:} Discussion above renders the adaptation of CLIP essential for effective AD. However, since CLIP’s embeddings are already optimized through extensive pretraining, it could be susceptible to overfitting to new dataset during adaptation. Overfitting convergence leads to minimized intra-class distinctions in the training data, often at the expense of the feature separability for effective generalization to unseen data.

To address this, \textit{a carefully controlled refinement} is crucial to preserve CLIP's generalization capabilities while enhancing its sensitivity to anomalies. 

\subsubsection{Overview of Our Solution}
Motivated by \cref{sec:current-challenges}, we propose \textbf{\Ourname(\ourname)} with improved anomaly awareness. As shown in \cref{fig:main},  \ourname is trained through a two-stage training strategy that sequentially adapts the semantic-rich text space and detail-focused visual space, with original CLIP parameters remaining frozen. In the first stage (see \cref{fig:main} (Top)), we incorporate Residual Adapters into the shallow layers of the text encoder, and the visual features from the fixed image encoder serve as a stable reference for optimization. A Disentangle Loss is purposed to enforce effective discrimination by ensuring independence between normal and anomaly embeddings. In the second stage, we integrate Residual Adapters into the shallow layers of the visual encoder to align patch-level features with the fixed, specially adapted texture features from the fixed text encoder (see in \cref{fig:main} (Bottom)). Ultimately, our \ourname succeeds in equipping CLIP with anomaly awareness across seen and unseen classes, as shown in \cref{fig:tsne} (bottom).
  
{\small
\begin{figure*}[!t]
    \centering
    \vspace{-2.5mm}
    \includegraphics[width=0.93\linewidth]{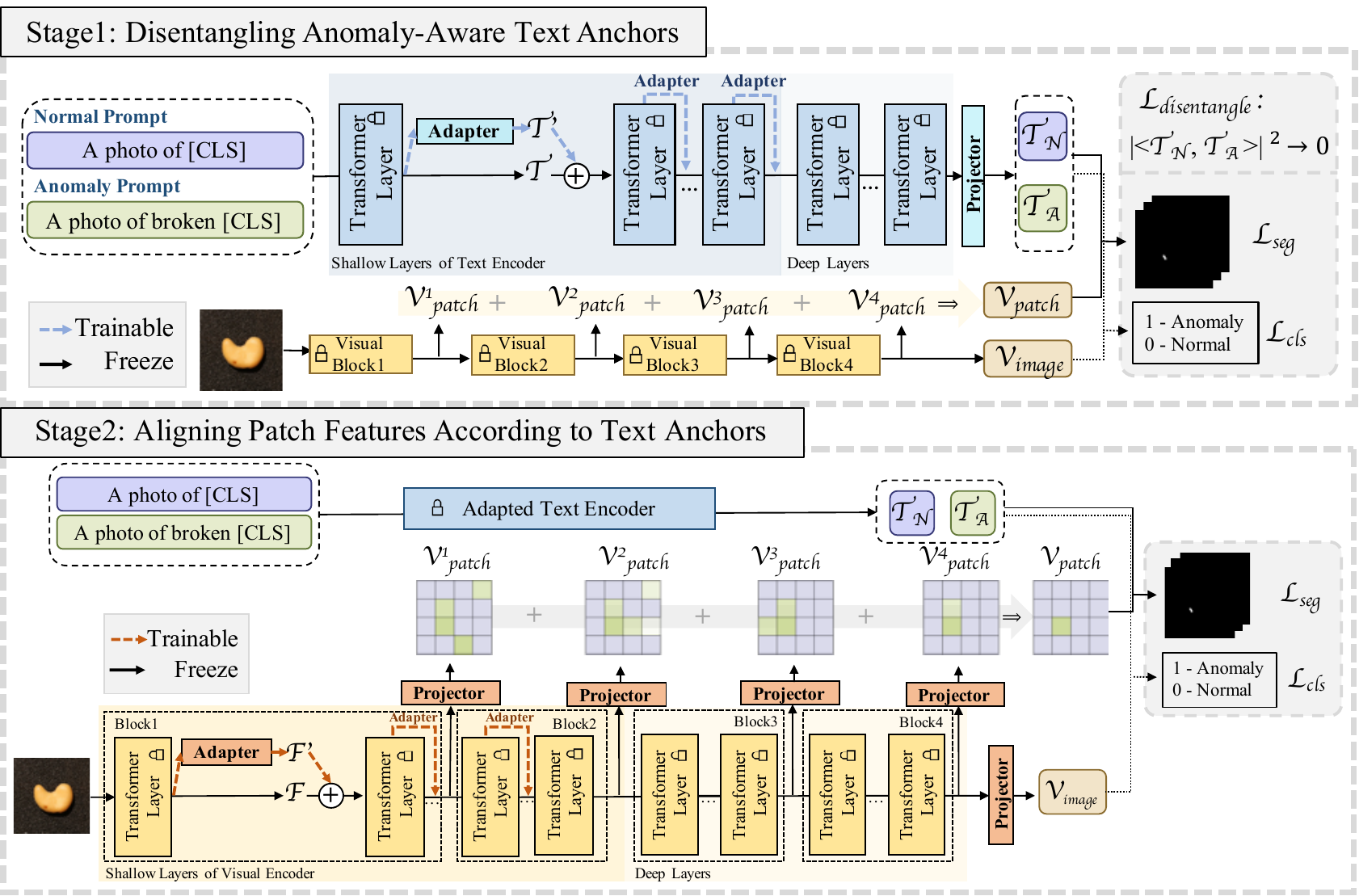}
    \vspace{-2mm}
    \caption{\textbf{The Two-Stage Training Pipeline of \Ourname.} In the first stage, the text encoder of \ourname is trained to identify anomaly-related semantics, helped by a disentangle loss. In the second stage, patch features are aligned with these text anchors. Both stages are achieved by the integration of Residual Adapters into the shallow layers of CLIP's backbone. This controlled adaptation enables CLIP to effectively distinguish anomalies, which forms our \Ourname.}
    \label{fig:main}
    \vspace{-2.5mm}
\end{figure*}
}

\subsection{\ourname with Two-Stage Adaptation Strategy}
\subsubsection{Residual Adapter}
To preserve CLIP's pre-trained knowledge while enabling targeted adaptation, we introduce lightweight Residual Adapters in the shallow layers (up to layer $K$) of both text and vision encoders.

The output feature $x^i\in \mathbb{R}^{N\times d}$ of CLIP's $i$-th ($i\leq K$) transformer layer is fed into the $i$-th adapter, outputting adapted feature $x_{residual}^i$, as shown in \cref{eq:adapter},
\begin{equation}
\begin{aligned}
  x_{residual}^i&=Norm(Act(W^ix^i)),
\end{aligned}
\label{eq:adapter}
\end{equation}
where $W^i\in \mathbb{R}^{d\times d}$ is the trainable linear weight of $i$-th adapter, $Act(\cdot)$ is an activation function, and $Norm(\cdot)$ is a normalizing function.
The original feature $x^i$ and the enhanced feature $x_{residual}^i$ are fused in a weighted manner, generating $x^i_{enhanced}$, the input to the next transformer layer, as shown in \cref{eq:residual}, 
\begin{equation}
\begin{aligned}
x^i_{enhanced}=\lambda\ x_{residual}^i + (1-\lambda)\ x^i,
\end{aligned}
  \label{eq:residual}
\end{equation}
where $\lambda$ is a hyper-parameter to control the 
residual ratio, adjusting the fusing degree of AD-specific knowledge for preserving the original CLIP's generalization ability and improved performance.

\subsubsection{Two-Stage Training Strategy}
\paragraph{Disentangling Anomaly-Aware Text Anchors:}
In the first stage, our objective is to learn anomaly-discriminative text anchors by adapting the text encoder while keeping the image encoder fixed. We incorporate Residual Adapters into the first $K_T$ layers of the CLIP text encoder, as illustrated in \cref{fig:main} (Top), and set the final projector in the text encoder to be learnable to facilitate improved alignment.

Using prompts designed to encapsulate both normal and anomalous semantics (as detailed in Appendix), text encoder generates corresponding high-level embeddings. The average embeddings of the normal and anomaly prompts serve as our initial text anchors, denoted as $T_N$ and $T_A\in \mathbb{R}^d$, respectively. These anchors are refined by being aligned with visual features extracted from an enhanced CLIP visual encoder, as \cite{li2023clip,zhou2023anomalyclip}. 
Alignment is conducted at both image and patch levels to incorporate both global and local semantics. By calculating the cosine similarity between these anchors and the image features $V_{image}\in \mathbb{R}^d$ or patch features $V_{patch}\in \mathbb{R}^{N\times d}$, as shown in \cref{eq:cos}, 
\begin{equation}
\begin{aligned}
p_{cls}&=CosSim(V_{image}, [T_N,T_A]), \\
p_{seg}^o &=CosSim(V_{patch}, [T_N,T_A]), 
\end{aligned}
  \label{eq:cos}
\end{equation}
where $[\cdot, \cdot]$ means concatenate operation, we obtain the classification prediction $p_{cls}\in \mathbb{R}^2$ and the segmentation prediction $p_{seg}^o\in\mathbb{R}^{N\times 2}$. The segmentation prediction $p_{seg}^o$ is then reshaped and upsampled to $p_{seg}\in \mathbb{R}^{H\times W \times 2}$ to align with the height $H$ and width $W$ of segmentation mask $S$. Following previous works~\cite{zhou2023anomalyclip, cao2025adaclip, huang2024adapting, chen2023april}, we compute the classification loss $\mathcal{L}_{cls}$ and segmentation loss $\mathcal{L}_{seg}$ to optimize parameters, as specified in \cref{eq:align}. Specifically, the classification loss is a binary cross-entropy that compares classification predictions with ground-truth labels $y$, and the segmentation loss is a combination of dice loss and focal loss applied to segmentation predictions and the anomaly segmentation mask $S$.
\begin{equation}
\begin{aligned}
&\mathcal{L}_{cls} =\text{BCE} (p_{cls}, y), \\
&\mathcal{L}_{seg} =\text{Dice} (p_{seg}, S)+\text{Focal} (p_{seg}, S), \\
&\mathcal{L}_{align} =\mathcal{L}_{cls}+\mathcal{L}_{seg}.
\end{aligned}
  \label{eq:align}
\end{equation}

To enhance the separation between normal and anomaly text embeddings, we introduce a Disentangle Loss encouraging orthogonality between $T_N$ and $T_A$ to minimize correlation, as in \cref{eq:dis}: 
\begin{equation}
\mathcal{L}_{dis}=|<T_N,T_A>|^2.
  \label{eq:dis}
\end{equation}
The Disentangle Loss $\mathcal{L}_{dis}$ is incorporated into the alignment loss $\mathcal{L}_{align}$ as a regularization term, weighted by a factor $\gamma$, which forms the total loss, as in \cref{eq:total}:
\begin{equation}
\mathcal{L}_{total}=\mathcal{L}_{align}+\gamma \mathcal{L}_{dis}.
  \label{eq:total}
\end{equation}

In this stage, the distinction between normal and anomaly semantics is embedded into CLIP's text encoder while its original object-recognition capability is preserved. \Cref{fig:tsne} indicates that this ability of anomaly-awareness is robust and generalizable to novel classes. 

\vspace{-3.3mm}
\paragraph{Aligning Patch Features According to Text Anchors:}
Anomaly-aware semantic anchors can facilitate the adaptation of patch features, thereby improving the effectiveness and generalizability of anomaly localization.
To achieve alignment between patch features and anchors from the previous stage, we introduce trainable Residual Adapters within the initial $K_I$ layers of the CLIP visual encoder.

Features with multi-granularities are utilized to enhance segmentation~\cite{zhou2023anomalyclip,jeong2023winclip,chen2023april}. Specifically, as shown in \cref{fig:main} (bottom), the intermediate output feature $F^i$ are extracted from four distinct granularities. These multi-granularity features are then projected to align with the channel of text anchors via a trainable projector $Proj_i(\cdot)$, yielding $V_{patch}^i$ at four distinct levels of granularity.
The aggregated output $V_{patch}$ is computed by summing individual $V_{patch}^i$ outputs, as in \cref{eq:agg}:
\begin{equation}
\begin{aligned}
    V_{patch}^i&=Proj_i(F^i),\ i\in \{1,2,3,4\} \\
    V_{patch}&=\sum^4_{i=1}V_{patch}^i.
\end{aligned}
  \label{eq:agg}
\end{equation}
The cosine similarity scores between the aggregated $V_{patch}$ and the text anchors are calculated to generate patch-level predictions as \cref{eq:cos}, resulting in the prediction maps.

During training, alignment is guided by the loss function defined in \cref{eq:align}, facilitating both global and local alignment. During inference, anomaly prediction maps and corresponding anomaly scores are derived by comparing the similarity scores of visual features against normal and anomaly text embeddings.
\section{Experiments}
\subsection{Experiment Setups}
\noindent\textbf{Datasets}
We evaluate our model on 11 widely used benchmarks, as previous AD works~\cite{zhou2023anomalyclip, huang2024adapting,cao2025adaclip,chen2023april,jeong2023winclip}, with distinct foreground objects spanning a variety of modalities, including photography, endoscopy, CT, MRI, and OCT. For the industrial domain, we use MVTec AD~\cite{bergmann2019mvtec}, VisA~\cite{zou2022spot}, BTAD~\cite{mishra2021vt} and MPDD~\cite{jezek2021deep}. For medical domain, we use brain MRI, liver CT and retina OCT from BMAD~\cite{bao2024bmadbenchmarksmedicalanomaly}, and four different colon polyp detection datasets with different views (CVC-ClinicDB~\cite{bernal2015wm}, CVC-ColonDB~\cite{bernal2012towards}, Kvasir-SEG~\cite{Pogorelov:2017:KMI:3083187.3083212} and CVC-300~\cite{vazquez2017benchmark}). Each dataset has both image-level labels and pixel-level masks for evaluation.

We train our model on a real-world industrial AD dataset - VisA~\cite{zou2022spot} - in which objects are different from other datasets. Results of VisA are obtained using MVTec-AD as the training dataset. To demonstrate adaptation efficiency, we conduct training under various data levels: 2-shot per class, 16-shot per class, 64-shot per class, and full-shot. The corresponding number of samples are randomly selected from each class, while maintaining a consistent 1:1 ratio between normal and anomaly samples.

\noindent\textbf{Metrics}
Following ~\cite{deng2022anomaly,you2022unified,defard2021padim,roth2022towards, huang2024adapting,cao2025adaclip,chen2023april,jeong2023winclip}, we use the Area Under the Receiver Operating Characteristic Curve (AUROC) as the metric. We compute AUROC at both the image and pixel levels to comprehensively assess the model's effectiveness in detecting and localizing anomalies.

\noindent\textbf{Implementation Details}
Following~\cite{cao2025adaclip,zhou2023anomalyclip,chen2023april,qu2024vcp}, we use OpenCLIP with the ViT-L/14 architecture as the backbone, and input images are resized to 518$\times$518. All parameters of CLIP remain frozen. We set $\lambda$ to 0.1, $K_T$ to 3, $K_I$ to 6, and $\gamma$ to 0.1. For multi-level feature extraction, we utilize outputs from the 6-th, 12-th, 18-th, and 24-th layers of the visual encoder to compose the overall output. For the first stage, we train the model for 5 epochs with a learning rate of $1\times 10^{-5}$. For the second stage, we continue training for 20 epochs, adjusting the learning rate to $5\times 10^{-4}$. Parameters are updated by Adam optimizers. All experiments are conducted on a single NVIDIA GeForce RTX 3090 GPU. More details are available in Appendix.

\subsection{Comparison with SOTA Methods}
\begin{table*}[t]
	\centering
    \resizebox{2.1\columnwidth}{!}{	
	\begin{tabular}{l|l|cccccc|cccc}
		\toprule
		\multicolumn{1}{l|}{\multirow{3}{*}{\!\!{Domain}\!\!}} & \multirow{2}{*}{{Dataset}} & CLIP$^{*}$ & WinCLIP$^{*}$  & VAND$^{*}$  & MVFA-AD & {\!\!\!\!\!\! AnomalyCLIP$^{*}$\!\!\!\!\!\!}& AdaCLIP & \multicolumn{4}{c}{\textbf{Ours}}\\ 
        \cline{3-8} \cline{9-12} 
        & & \footnotesize{OpenCLIP} & \small{CVPR 2023} & \small{CVPRw 2023} & \small{CVPR 2024} & \small{ICLR2024} &\small{ECCV2024}&\multicolumn{4}{c}{-}\\
        \cline{2-2} \cline{3-8} \cline{9-12} 
        & \small{Available training shots} & - & - & full & full & full & full & 2 & 16 & 64 & full \\
        \midrule
        \multirow{3}{*}{Industrial}
        & BTAD & 30.6 & 32.8 & 91.1 & 90.1 &  93.3 & 90.8 & 92.8 & \rd 94.4 & \nd 96.5 & \fs 97.0\\
        & MPDD & 62.1 & 95.2 & 94.9 & 94.5 & 96.2 &  96.6 &  96.3 & \nd 96.5 &  \rd 96.3  & \fs 96.7 \\
        &MVTec-AD & 38.4 & 85.1 & 87.6 & 84.9 &  91.1 & 89.9 & 91.0 & \rd 91.2 & \nd 91.6 & \fs 91.9 \\
        &\textit{VisA} & 46.6 & 79.6 & 94.2 & 93.4 &  \nd 95.4 & \fs 95.5 & 93.4 & 93.8 & 94.0 & \fs 95.5 \\
        \midrule
        \multirow{6}{*}{Medical}&Brain MRI &68.3 & 86.0 & 94.5 &95.6 &  96.2 & 93.9 & \rd 96.3 &  \nd 96.4 & \fs 96.5 & 95.5 \\
        & Liver CT & 90.5 & 96.2 & 95.6 & 96.8 & 93.9 & 94.5 &  97.3 & \rd 97.7 & \nd 97.7 & \fs 97.8  \\
        & Retina OCT & 21.3 & 80.6 & 88.5 & 90.9 & 92.6 & 88.5 &  94.2 & \nd 95.1 & \rd 94.4 & \fs 95.5 \\
        & ColonDB & 49.5 & 51.2 & 78.2 & 78.4 &  82.9 & 80.0 & \rd 83.9 &  83.5 & \fs 84.7 & \nd 84.0 \\
        & ClinicDB & 47.5 & 70.3 & 85.1 & 83.9 & 85.0 & 85.9 & \nd 89.2 &  87.6 & \rd 87.8 & \fs 89.9 \\
        & Kvasir & 44.6 & 69.7 & 80.3 &  81.9 &  81.9 & \nd 86.4 & 82.1 & 84.6 & \rd 85.2 & \fs 87.2 \\
        & CVC-300 & 49.9 & - & 92.8 & 82.6 & 95.4 & 92.9 & 96.0 & \fs 97.4 & \rd 96.0 & \nd 96.4 \\
        \midrule 
        \multicolumn{2}{c|}{Average} & 49.9 & 74.7 & 89.3 & 88.5 &  91.3 & 90.4 & 92.0 &  \rd 92.6 & \nd 92.8 & \fs 93.4 \\
		\bottomrule
	\end{tabular}
    }
    \vspace{-1mm}
	\caption{Pixel-level AUROC of zero-shot AD methods in Industrial and Medical domains. Method sources and the number of shots used for training are noted. Results of methods with $^*$ are copied from the papers or inferred from official weight.
    Best results are highlighted as \colorbox{colorFst}{\bf \!first\!}, \colorbox{colorSnd}{\!second\!} and \colorbox{colorTrd}{\!third\!}.}
	\label{tb:pixel}
    \vspace{-1.5mm}
\end{table*}
\begin{table*}[t]
	\centering
    \resizebox{2.1\columnwidth}{!}{	
	\begin{tabular}{l|l|ccccc|cccc}
		\toprule
		\multicolumn{1}{l|}{\multirow{3}{*}{\!\!{Domain}\!\!}} & \multirow{2}{*}{{Dataset}} & {\!\!CLIP\&VAND$^{*}$\!\!} & {\!\!WinCLIP$^{*}$\!\!} & {\!\!MVFA-AD\!\!\!} & {\!\! AnomalyCLIP$^{*}$\!\!}  & {\!\!AdaCLIP \!\!} & \multicolumn{4}{c}{\textbf{Ours}}\\ 
        \cline{3-7} \cline{8-11} 
        & & \small{OpenCLIP} & \small{CVPR 2023} & \small{CVPR 2024} & \small{ICLR2024} &\small{ECCV2024} &\multicolumn{4}{c}{-}\\
        \cline{2-2}\cline{3-7} \cline{8-11} 
        & \small{Available training shots} & - & - & full & full & full & 2 & 16 & 64 & full \\
        \midrule
        \multirow{3}{*}{Industrial}
        & BTAD & 73.6 & 68.2 & \rd 94.3 &85.3 & 90.9 & 88.0 &  90.9 &\nd 94.7 & \fs 94.8\\
        & MPDD &  73.0 & 63.6 & 70.9 &  73.7 &  72.1 & 63.6 & \fs 78.3 &\nd 75.7 &\rd 75.1 \\
        & MVTec-AD & 86.1 & \nd 91.8 & 86.6 & \rd 90.9 & 90.0 & 85.9 & 89.7 & \fs 92.0 &  90.5 \\
        & \textit{VisA} & 66.4 &  78.0 & 76.5 &  82.1 & \nd 84.3 & 78.4 & \rd 84.0 & 84.1 & \fs 84.6 \\
        \midrule
        \multirow{3}{*}{Medical}&Brain MRI & 58.8 & 66.5 & 70.9 & \rd 83.3 &  80.2 & \fs 84.3 & 80.4 & \nd 83.4 & 80.2  \\
        & Liver CT & 54.7 & 64.2 & 63.0 &61.6 &  64.2 & \nd 69.4 & 68.1 & \rd 69.2 & \fs 69.7 \\
        & Retina OCT & 65.6 & 42.5 & 77.3 & 75.7 & \rd 82.7 & 77.4 &  81.0 & \fs 82.9 & \nd 82.7 \\
        \midrule
        \multicolumn{2}{c|}{Average}  & 68.3 & 67.8 & 77.1 &  78.4 &  80.6 & 78.1 & \rd 81.8 & \fs 83.1 & \nd 82.5 \\
		\bottomrule
	\end{tabular}
    }
    \vspace{-1mm}
	\caption{Image-level AUROC of zero-shot AD methods in Industrial and Medical domains. Method sources and the number of shots used for training are noted. Results of methods with $^*$ are copied from the papers or inferred from official weight.
    Best results are highlighted as \colorbox{colorFst}{\bf \!first\!}, \colorbox{colorSnd}{\!second\!} and \colorbox{colorTrd}{\!third\!}.}
	\label{tb:image}
    \vspace{-2.5mm}
\end{table*}
We compare our method against CLIP and several recent SOTA models. Among them, WinCLIP~\cite{jeong2023winclip}, VAND~\cite{chen2023april} and MVFA-AD~\cite{huang2024adapting} use original CLIP text encoder, and AnomalyCLIP~\cite{zhou2023anomalyclip} and AdaCLIP~\cite{cao2025adaclip} incorporate learnable prompts. To ensure a fair comparison, we re-train models that are originally trained on different datasets to match the dataset settings of other approaches (detailed in Appendix).

Quantitative results are presented in \cref{tb:pixel} and \cref{tb:image}. Although adapting only the patch feature with original text embeddings has made progress in AD, the superior performance of \ourname highlights its effective disentanglement of anomaly-discriminative semantics, leading to further progress. Notably, even in data-limited situations, our method consistently demonstrates top performance. At the pixel level, with only 2 shots per class used for training, our method achieves improved average zero-shot performance compared to previous methods. With the full dataset, we set a new pixel-level SOTA with an AUROC of 93.4\%. At the image level, our method is competitive with just 2 shots for training and establishes a new SOTA of 83.1\% with 64 shots per class.

Unlike previous methods, our approach does not rely heavily on data resources to achieve top-tier performance. Comparison under different levels of data available, as shown in \cref{fig:plot}, reveals that our approach consistently outperforms other methods in general. Even with limited data, our model reaches competitive results, while other methods display signs of underfitting. As data increases, our method maintains its lead, establishing a new SOTA at both pixel and image levels. 
{\small 
\begin{figure}[!t]
    \centering
    \includegraphics[width=0.85\linewidth]{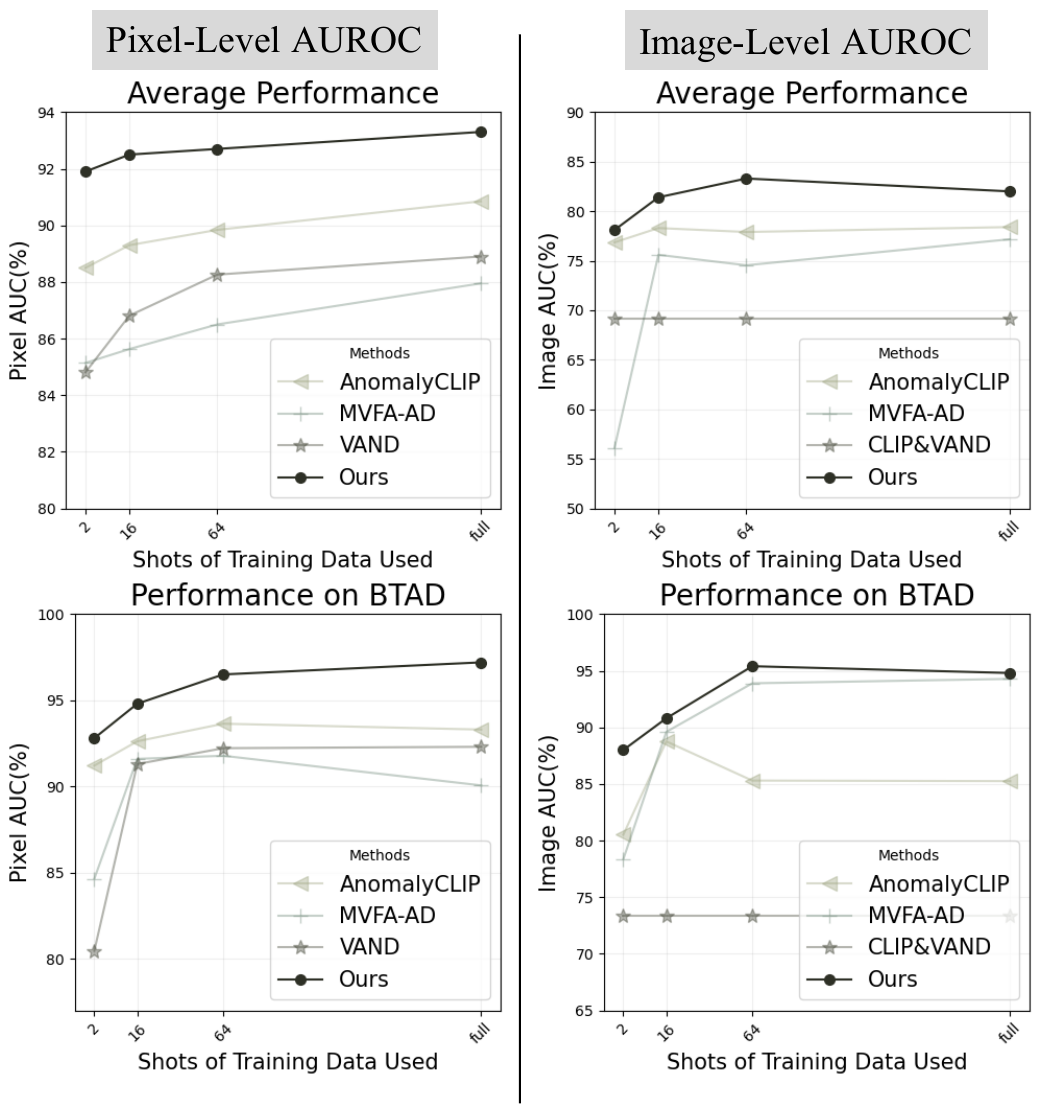}
    \vspace{-2mm}
    \caption{\textbf{Average Results (Top) and Results on BTAD (Bottom) of Different methods Trained on 2-, 16-, 64-shot per Class and Full Data of VisA.} Our method shows high fitting efficiency, achieving strong results across all data scales. }
    \vspace{-2mm}
    \label{fig:plot}
\end{figure}
}

\subsection{Visualization}
To illustrate the alignment intuitively, we present visualization examples in \cref{fig:vis} with original configuration for previous works. Although previous methods with can detect anomalous regions, our \ourname demonstrates fewer false-negative predictions in both industrial and medical domains, accurately highlighting the correct anomaly regions. 

{\small 
\begin{figure}[!t]
    \centering
    \includegraphics[width=0.95\linewidth]{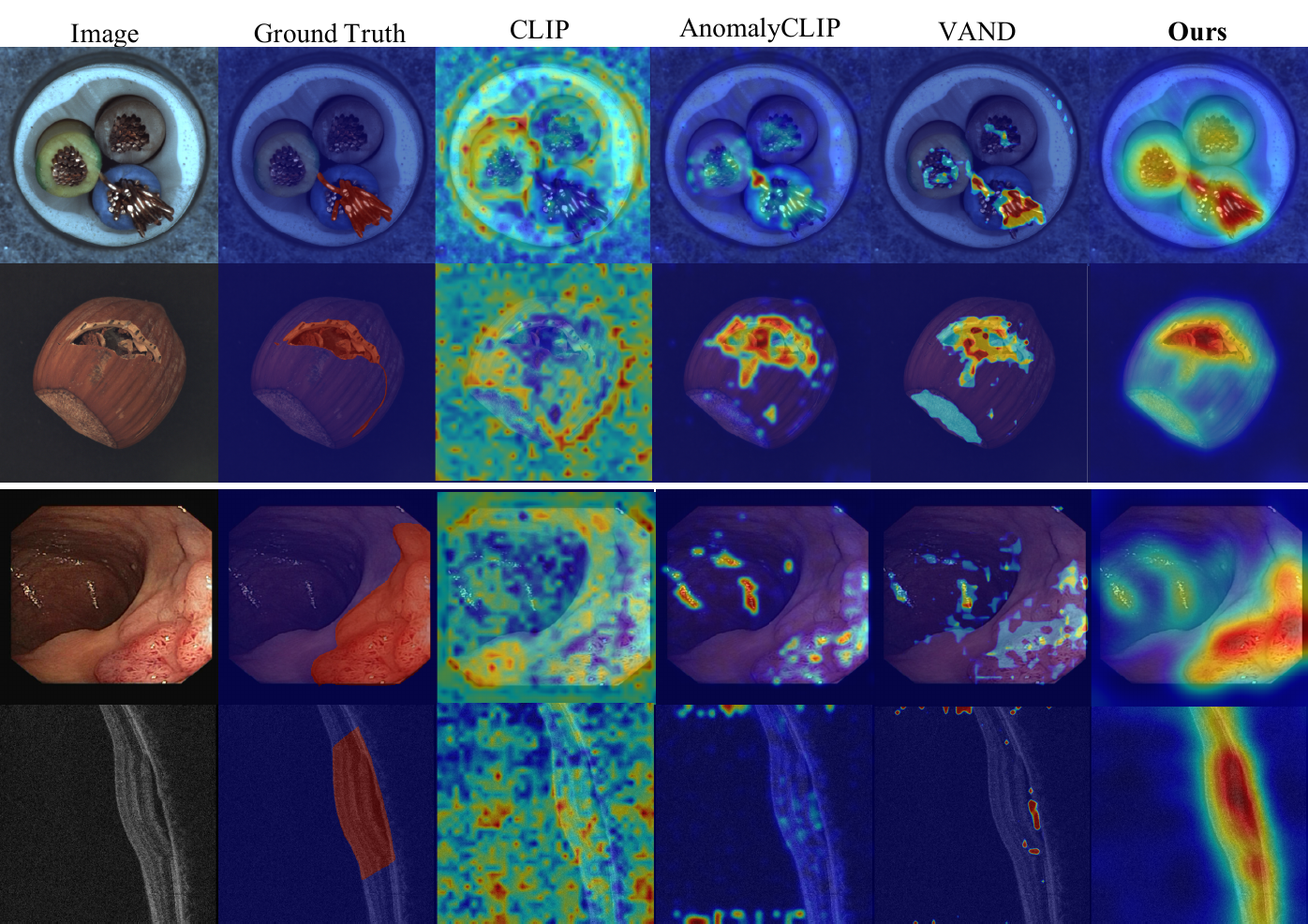}
    \vspace{-1mm}
    \caption{\textbf{Visualization of Anomaly Localization Results of Original CLIP~\cite{radford2021learning}, AnomalyCLIP~\cite{zhou2023anomalyclip}, VAND~\cite{chen2023april} and our \ourname.} Compared to previous methods, \ourname demonstrates more reliable prediction capabilities in localizing anomaly.}
    \vspace{-3mm}
    \label{fig:vis}
\end{figure}
}

\subsection{Ablations Analysis}
\begin{table}[t]
	\centering
    \resizebox{\columnwidth}{!}{	
	\begin{tabular}{lcl|cc}
		\toprule
		\multicolumn{3}{c|}{\multirow{2}{*}{\!\!{Method}\!\!}} & \multicolumn{2}{c}{\!\!{Avg. AUROC}\!\!} \\
        \cline{4-5}
        & & & Pixel-Level & Image-Level \\
        \midrule
        \multicolumn{3}{c|}{CLIP} & 50.3 & 69.3 \\
        \midrule
        \multirow{3}{*}{Image} & 1. & + Linear Proj. \textit{(VAND~\cite{chen2023april})} & 88.9 & 69.3 \\
        & 2. & + Adapter & 48.9\color{gray}{(-40.0)} & 53.4\color{gray}{(-15.9)} \\
         & 3. & \textbf{+ Residual Adapter} & 91.3\color{gray}{(+2.4)} & 80.7\color{gray}{(+11.4)} \\
        \midrule
        \multirow{2}{*}{Text} & 4. & \textbf{+ Residual Adapter} & 92.1\color{gray}{(+3.2)} & 82.6\color{gray}{(+13.3)} \\
        & 5. & \textbf{+ Disentangle Loss} & 92.7\color{gray}{(+3.8)} & 83.3\color{gray}{(+14.0)} \\
		\bottomrule
	\end{tabular}
    }
    \vspace{-1mm}
	\caption{\textbf{Ablation Study of Our Training Strategy with VisA-Trained 64-Shot Setup.} Our contributions are \textbf{bold}. While VAND uses linear projectors to improve AD performance, incorporating Residual Adapters further refines patch feature adaptation. Moreover, integrating our Disentangle Loss yields the best overall results.}
    \vspace{-3.8mm}
	\label{tb:ablation}
\end{table}
We conduct thorough ablation experiments of our refinement of both visual and text space, as shown in \cref{tb:ablation} and \cref{fig:one-stage}. The second row in \cref{tb:ablation}, which mirrors the structure of VAND~\cite{chen2023april}, serves as our baseline. 

\vspace{1.5mm}
\noindent{\textbf{Image Space:}} As shown in \cref{tb:ablation} line ``2.'', inserting the vallina linear adapter into transformer layers results in a significant decline in zero-shot performance, indicating the damage of the original generalization ability of CLIP. Incorporating our Residual Adapters mitigates this issue (shown in line ``3.''), enhancing performance while preserving original information stored in CLIP. 

\vspace{1.5mm}
\noindent{\textbf{Text Space:}} The last two rows in \cref{tb:ablation} highlight the impact of our approach in equipping CLIP's encoder with anomaly-aware semantics. Line ``4.'' validates that, with \ourname, the model's ability to discriminate anomalies further improves, as the \ourname's text encoder provides a more precise semantic foundation. Adding Disentangle Loss leads to an additional improvement (shown in Line ``5.''), especially at image-level, validating the necessity of independence between normal and anomaly anchors. These results underscore the crucial role of text space refinement in improved anomaly localization and classification.

\vspace{1.5mm}
\noindent{\textbf{Two-Stage Training:}}
To validate the necessity of two-stage training, we adapt both text and image encoders together within one stage (also adopted by AdaCLIP). As shown in ~\cref{fig:one-stage}, one-stage model can easily exaggerate anomaly semantics and forget class information embedded in CLIP, damaging the model's generalization ability. The two-stage training strategy allows controlled adaptation, preserving CLIP's class-relevant knowledge in one end while adapting the other, as shown in ~\cref{fig:tsne}.
{\small 
\begin{figure}[!t]
    \centering
    \includegraphics[width=\linewidth]{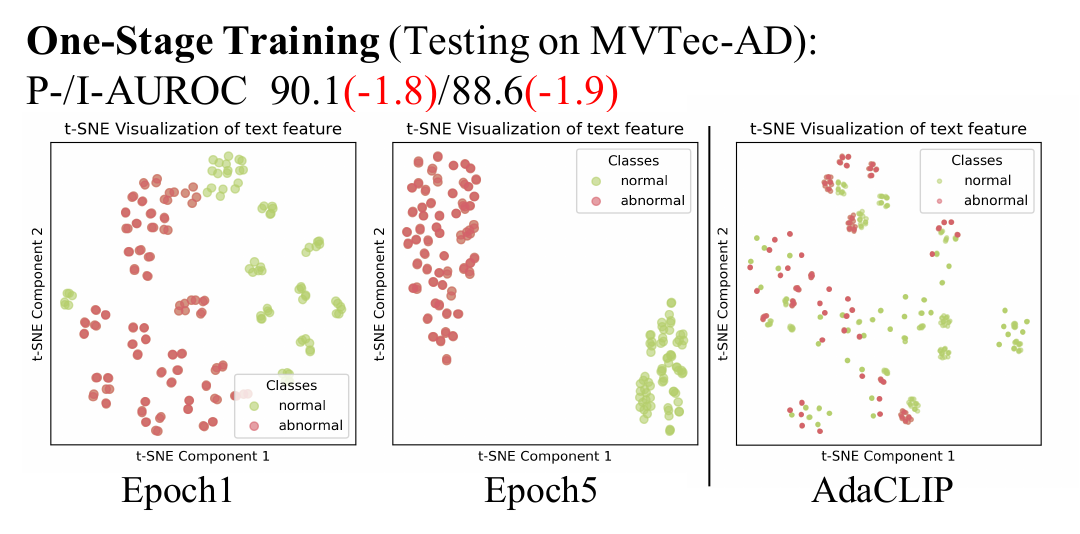}
    \vspace{-6mm}
    \caption{\textbf{Visualization of Text Space from One-Stage Training and from AdaCLIP.} During one-stage training, class information collapses easily, leading to damaged zero-shot performance.}
    \vspace{-2mm}
    \label{fig:one-stage}
\end{figure}
}

\section{Conclusion and Discussion}

To our knowledge, this is the first work to explicitly analyze the intrinsic Anomaly Unawareness problem in CLIP. To tackle this issue, we propose a simple yet effective two-stage training strategy to embed anomaly-aware information into CLIP, enabling clear disentanglement of anomaly representations across both seen and novel classes. By leveraging residual adapters, our method preserves CLIP’s strong generalization ability, achieving outstanding zero-shot performance across multiple datasets.

Our adapted \ourname, developed through this two-stage adaptation strategy, reveals the potential of refining CLIP’s feature space for improved performance in downstream applications. 
Beyond addressing anomaly unawareness, our work also provides a potential foundation for tackling other ``unawareness'' issues within CLIP. These may include limitations in context-awareness or specificity to domain-relevant nuances, suggesting further applications of our method in expanding CLIP’s adaptability across diverse tasks. Additionally, we observe signs of overfitting with full-shot training, suggesting potential saturation during CLIP adaptation and warranting further investigation.

\section*{Acknowledgement}
This work is supported by Natural Science Foundation of China under Grant 62271465, Suzhou Basic Research Program under Grant SYG202338, Open Fund Project of Guangdong Academy of Medical Sciences, China (No. YKY-KF202206), and Jiangsu Province Science Foundation for Youths (NO. BK20240464).


{
\small
\bibliographystyle{ieeenat_fullname}
\bibliography{main}
}

\end{document}